\documentclass[10pt,twocolumn,letterpaper]{article}

\usepackage{iccv}
\usepackage{times}
\usepackage{epsfig}
\usepackage{graphicx}
\usepackage{amsmath}
\usepackage{amssymb}

\usepackage[breaklinks=true,bookmarks=false]{hyperref}

\iccvfinalcopy 

\def\iccvPaperID{3} 

\begin{document}

\title{Multi-level Domain Adaptive learning for Cross-Domain Detection}
\author{Rongchang Xie$^{1*}$, Fei Yu$^{1*}$, Jiachao Wang$^{1}$, Yizhou Wang$^{2,4,5}$, Li Zhang$^{1,3}$\\
\tiny \\
	\normalsize $^{1}$Center for Data Science, Peking University  ~~~~ $^{2}$Computer Science Dept., Peking University\\
	\normalsize $^{3}$Center for Data Science in Health and Medicine, Peking University\\
	\normalsize  $^{4}$Peng Cheng Lab ~~~~~~    $^{5}$ Deepwise AI Lab   ~~~~~~~    $^*$ These authors contributed equally\\\
	{\tt\small \{rongchangxie, yufei1900, wangjiachao, yizhou.wang,  zhangli\_pku\}@pku.edu.cn}
}
\maketitle

\begin{abstract}
In recent years, object detection has shown impressive results using supervised deep learning, but it remains challenging in a cross-domain environment. The variations of illumination, style, scale, and appearance in different domains can seriously affect the performance of detection models. Previous works use adversarial training to align global features across the domain shift and to achieve image information transfer. However, such methods do not effectively match the distribution of local features, resulting in limited improvement in cross-domain object detection. To solve this problem, we propose a multi-level domain adaptive model to simultaneously align the distributions of local-level features and global-level features. We evaluate our method with multiple experiments, including adverse weather adaptation, synthetic data adaptation, and cross camera adaptation. In most object categories, the proposed method achieves superior performance against state-of-the-art techniques, which demonstrates the effectiveness and robustness of our method.
\end{abstract}
\section{Introduction}
\begin{figure}[t]
	\begin{center}
		\includegraphics[width=1\linewidth]{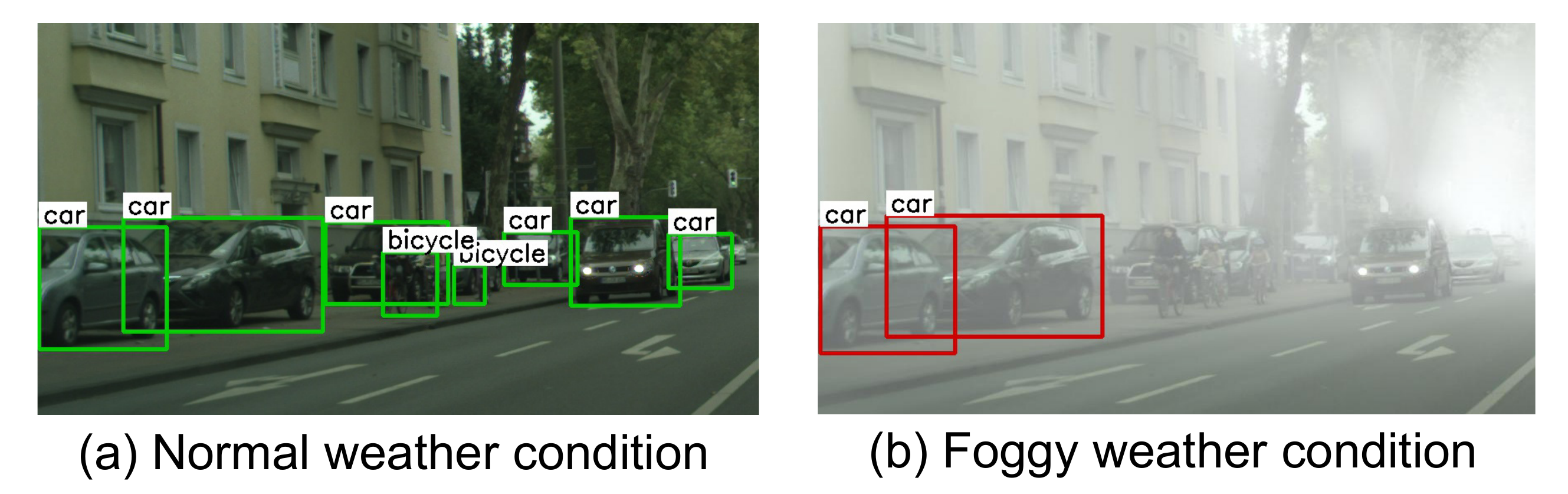}	
	\end{center}
	\caption{Illustration of domain shift. (a) Detection result of supervised training on \textit{Cityscapes}. (b) Detection result on \textit{Foggy Cityscapes} using the model trained on \textit{Cityscapes}. The weather conditions cause great performance drop.}
	\label{fig:domain_shift}
\end{figure}

With the rapid development of convolutional neural networks (CNN) in recent years, many major breakthroughs have been made in the field of object detection \cite{dai2016r, he2017mask, redmon2018yolov3, ren2015faster}. Detection models are getting faster, more reliable, and more accurate. However, domain shift remains one of the major challenges in this area. For example, as shown in Figure \ref{fig:domain_shift}, models trained with normal weather images are unable to effectively detect objects in foggy weather. This is because the domain shift causes significant differences in the features extracted from the two types of data, making it impossible to simply apply the model trained on the source domain directly to the unlabeled target domain.

Although collecting more data for training may alleviate this problem, it is non-trivial because the annotations in object detection are particularly burdensome. To tackle the domain shift problem without introducing additional annotations, many researchers propose various domain adaptation methods to transfer the knowledge of the label-rich domain to the label-poor domain \cite{ganin2016domain, hoffman2017cycada, liu2017unsupervised, tzeng2017adversarial}. Such methods use adversarial learning to minimize the $\mathcal{H}$-divergence between the source and the target domains, searching for an appropriate feature space in which the distribution of the source and target objects is well aligned. Therefore, the model can extract domain-independent features of the objects and correctly detecting them via knowledge transfer without requiring additional annotations.

The key to achieving such a goal is how to measure the learned features in different domains and examine whether they are consistent. For example, Domain-Adversarial Neural Network (DANN) \cite{ganin2015unsupervised} uses a domain classifier to measure the features and achieves end-to-end adversarial learning by reversing the gradient from the subsequent level to the preceding layers, thereby improving the consistency of feature distributions in different domains. Domain Adaptive (DA) Faster R-CNN \cite{chen2018domain} extends this idea to object detection, matching features in the image stage and instance stage. In a more recent work, the authors demonstrate that global matching in DA Faster R-CNN may only address small domain shifts, but the detection accuracy related to large domain shifts may decrease \cite{saito2019strong}. Therefore, they propose a Strong-Weak DA Faster R-CNN that combines weak global alignment with strong local alignment.

In this work, we expect to optimize the original DA Faster R-CNN model to achieve more accurate cross-domain object detection without using additional annotations. One of the main problems with existing methods is that light-weighted domain classifiers cannot form effective adversarial learning with complex Faster R-CNNs. That is, Faster R-CNN can easily deceive the domain classifier, so that the feature alignment is highly possible to be ineffective. Inspired by the Strong-Weak DA Faster R-CNN~\cite{saito2019strong}, we propose a Multi-level Domain Adaptive Faster R-CNN. Our model has two advantages: First, we use different domain classifiers to supervise the feature alignments from multiple scales; Second, more domain classifiers enhance the model's discriminating ability and optimize overall adversarial training. Experiments have shown that aligning the feature distributions of intermediate layers can also alleviate covariate shift and achieve better domain adaptation. Furthermore, our model also follows the conclusion in \cite{saito2019strong} that local alignment should be stronger than global alignment. Because during the backpropagation, the lower feature extractors in Faster R-CNN are getting the reversal gradient from all subsequent domain classifiers, which means it should maintain stronger ability of feature alignment to deceive more domain classifiers. For higher levels, the need for this ability will be appropriately weakened. 

We evaluate our approach on several datasets including \textit{Cityscapes} \cite{cordts2016cityscapes}, \textit{KITTI} \cite{geiger2013vision}, \textit{SIM 10k} \cite{johnson2017driving}. The qualitative and quantitative results demonstrate the effectiveness of our method for addressing the domain shift problem. Furthermore, the multiple domain classifiers are only used for model training and not for inference, which won't impact the inference efficiency.

\section{Related Work}
\paragraph{Domain adaptation}
Domain adaptation is a technique that adapts a model trained in one domain to another. Many related works try to define and minimize the distance of feature distributions between the data from different domains \cite{ganin2015unsupervised,long2015learning,ren2018cross,saito2018maximum, Tzeng_2017_CVPR,tzeng2014deep,zhang2018collaborative}. For example, deep domain confusion (DDC) model \cite{tzeng2014deep} explores invariant representations between different domains by minimizing the maximum mean discrepancy (MMD) of feature distributions. Long \etal propose to adapt all task-specific layers and explore multiple kernel variants of MMD \cite{long2015learning}. Ganin and Lempitsky report using the adversarial learning to achieve domain adaptation and learning the distance with the discriminator \cite{ganin2015unsupervised}.  Most of the mentioned works above are designed for classification or segmentation.

Huang \etal propose that aligning the distributions of activations of intermediate layers can alleviate the covariate shift \cite{huang2018domain}. This idea is similar to our work partly. However, instead of using a least squares generative adversarial network(LSGAN) \cite{mao2017least} loss to align distributions for semantic segmentation, we use multi-level image patch loss for object detection.

\paragraph{Domain adaptation for object detection}
Although domain adaptation has been studied for a long time in classification tasks, its application in object detection is still in its early stages. Chen \etal propose to align both image's features and instance's features to achieve cross-domain object detection \cite{chen2018domain}. Inoue \etal address cross-domain weakly supervised object detection using domain-transfer and pseudo labeling \cite{inoue2018cross}. More recently, Kim \etal use domain diversification and multi-domain-invariant representation learning to address the source-biased problem \cite{kim2019diversify}. Saito \etal propose global-weak alignment that puts less emphasis on aligning images that are globally dissimilar \cite{saito2019strong}. Zhu \etal focuses on mining the discriminative regions which are directly related to object detection and aligning them across different domains \cite{zhu2019adapting}.

\section{Method}

\begin{figure*}[!h]
	\begin{center}
		\includegraphics[width=0.88\linewidth]{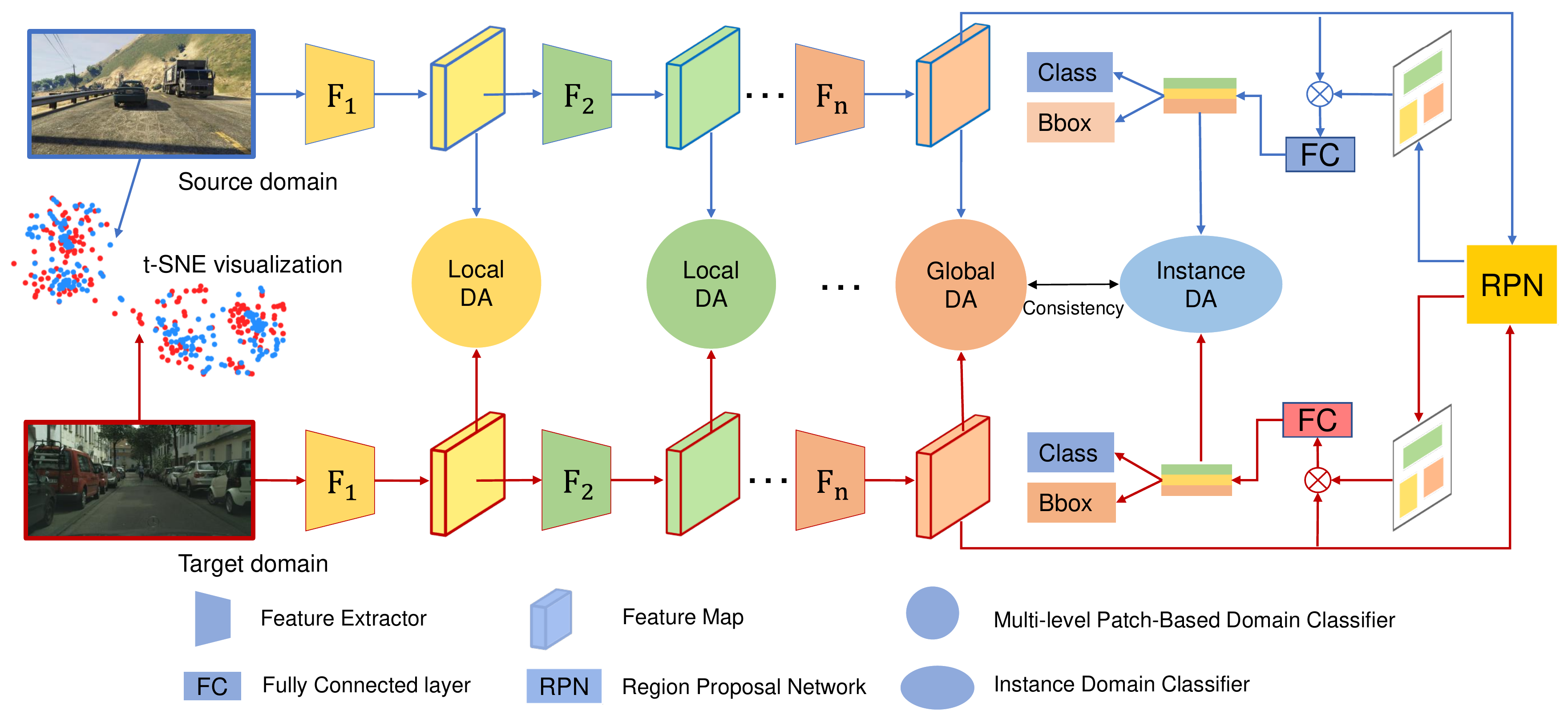}
	\end{center}
	\caption{The Illustration of proposed network. Our method uses multi-level domain classifiers to get a better alignment of feature distributions (see the t-SNE visualization). All domain classifiers utilize GRL layers to achieve adversarial learning.}
	\label{fig:network}
\end{figure*}

\subsection{Preliminaries}
Our work adopts the main idea of \textbf{Domain Adaptive Faster R-CNN (DA model)} \cite{chen2018domain}, which contains two major parts: 1. Image-Stage Adaptation; 2. Instance-Stage Adaptation.

\paragraph{Image-Stage Adaptation}  A domain classifier is used to predict the domain label for each image patch, which reduces the image-stage shift, such as image style, scale, \textit{etc.} The loss of image-stage adaptation can be formatted as,

\begin{equation}
\mathcal{L}_{i m g}=-\sum_{i, u, v}\left[D_{i} \log p_{i}^{u, v}+\left(1-D_{i}\right) \log \left(1-p_{i}^{u, v}\right)\right],
\label{eq_img}
\end{equation}

where $D_i$ denotes the domain label of the $i$-th image. And $p_{i}^{u, v}$ represents the probability that the pixel at $(u,v)$ on the final feature map belongs to the target domain. For each image patch, all corresponding activations on the feature maps will be classified. so we call this \textbf{patch-based domain adaptive loss}.

\paragraph{Instance-Stage Adaptation}
The instance-stage adaptation loss is defined as,
\begin{equation}
\mathcal{L}_{i n s}=-\sum_{i, j}\left[D_{i} \log p_{i, j}+\left(1-D_{i}\right) \log \left(1-p_{i, j}\right)\right],
\label{eq_ins}
\end{equation}

where $p_{i, j}$ represents the probability that the $j$-th region proposal in the $i$-th image is from target domain.

\subsection{Proposed Multi-level DA model}
\label{sec.method}

The \textbf{image-stage adaptation} of DA model uses the feature map after the last convolutional layer to align the global feature distribution of different domains. However, such a setting has two limitations. First, the model ignores the alignment of local features, making certain domain-sensitive local features weaken the generalization ability of the adaptive model. Second, single adaptation (one domain classifier) is difficult to cancel the data bias between the source domain and the target domain, because there are other non-transferable layers. \cite{long2015learning}.

\paragraph{Multi-level Patch-based Loss} To solve the aforementioned problems and to improve the cross-domain detection, we adopt the idea of layer-wise adaptation \cite{huang2018domain, long2015learning}.

We extract the output feature maps of multiple intermediate layers in the convolutional network, and build multiple corresponding image domain classifiers to supervise the feature alignments in the intermediate layers. The multiple level loss can be written as,

\begin{equation}
\begin{split}
\mathcal{L}_{multi}=-\sum_{i,k, u, v}[D_{i} \log f_k(\Phi_{i,k}^{u, v})+ \\
(1-D_{i}) \log(1-f_k(\Phi_{i,k}^{u, v}  ))],
\label{eq_mul}
\end{split}
\end{equation}

where $\Phi_{i,k}^{u, v}$ denotes the activation located at $(u, v)$ of the feature map of the $i$-th image after the $k$-th layers, and $f_k$ is its corresponding domain classifier. The multi-level domain adaptive components guarantee that the distributions of intermediate features in two domains are matched and enhance the robustness of the adaptive model. It is worth noting that the number of intermediate layers used in domain adaptation can vary in different datasets. For simplicity, the weight factors of all layers are set equal.

\paragraph{Total Objective Function}
We use Faster R-CNN \cite{ren2015faster} as our detection model. The instance-stage loss and consistency loss (detailed in \cite{chen2018domain}) are also adopted in our work. The overall objective is the summation of detection loss and adaptation loss. 
 
\begin{equation}
L=L_{d e t}+\lambda(L_{multi}+L_{ins}+L_{cst}),
\end{equation}

where $\lambda$ is a weight factor to balance detection loss and adaptation loss. The detection loss is composed of the loss of Region Proposal Network(RPN) and the loss of R-CNN. The total objective $L$ can be optimized by the standard SGD algorithm. 
 
As shown in Figure \ref{fig:network}, the main network architecture of our model contains a detection part (Faster R-CNN) and an adaptation part (local DA, global DA, and instance DA). In the training phase, the detection part is trained by labeled samples from source domain and the adaptation part is trained by samples from both source and target domains. We then use GRL layers \cite{ganin2015unsupervised} to reverse the sign of gradient, which encourages base network to learn domain-insensitive features. In the inference phase, only detection part will be used to predict the bounding boxes and classes of the objects. Thus, it will not lose time efficiency.

\begin{figure*}[!h]
	\begin{center}
		\includegraphics[width=0.95\linewidth]{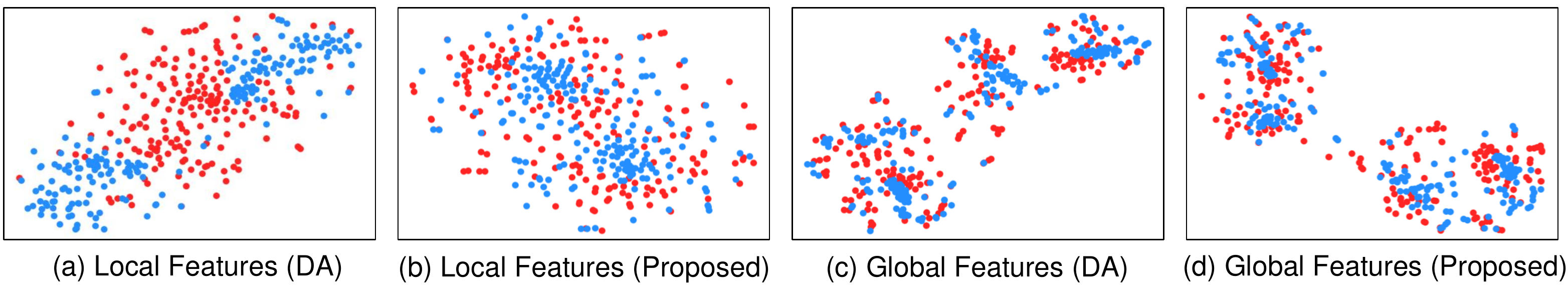}
	\end{center}
	\caption{Visualization of image features at different levels using the t-SNE algorithm: (a) Local features from DA model (b) Local features from our Multi-level model (c) Global features from DA model (d) Global features from our Multi-level DA model. Each point represents feature of an image patch. The red is from source domain and the blue is from target domain.}
	\label{fig:img_vis}
\end{figure*}

\section{Experiments and Results}
In order to evaluate the effectiveness of our approach, we follow \cite{chen2018domain} to perform three experiments: \textbf{1) Adverse Weather Adaptation.} In this experiment, we aim to adapt networks from detecting objects in normal weather to that in foggy weather. \textbf{2) Synthetic Data Adaptation.} In this experiment, we aim to adapt networks for the data from video games to the data from real world. \textbf{3) Cross Camera Adaptation.} In this experiment, we aim to adapt networks for the photos under different camera setups. In addition, we evaluate the visualization of feature distribution to support our claim that adding multiple domain classifiers can enhance the model's overall discriminating ability and achieve more appropriate alignments.

\subsection{Experiment details}
In all experiments, only the source training data are provided with annotations. We set the shorter side of the image to 600 pixels. The VGG-16 model \cite{simonyan2014very} is pretrained on ImageNet and is used as the network backbone of our model. Because the first four convolutional layers of VGG are fixed in traning, we distribute the discriminators at equal intervals from the fifth layer to the final convolutional layer (e.g. 5th/13, 9th/13, 13th/13).  The network is then finetuned for 6 epochs with a learning rate of 0.002 and for another 4 epochs with a learning rate of 0.0002. We also use weight decay and momentum, which are set as 0.0005 and 0.9. During the training process, we flip the images for data augmentation and feed two images from the different domains into the network in every iteration. To evaluate the proposed method, we report mean average precision (mAP) with a threshold of 0.5 on the last epoch. Without specific notation, we set $\lambda=0.1$. 

\subsection{Adverse Weather Adaptation}

\begin{table*}
	\begin{center}
		\begin{tabular}{|l|c|cccccccc|c|}
			\hline
			Method  & Backbone & Person & Rider & Car & Truck & Bus & Train & Motorcycle & Bicycle & Mean AP \\
			\hline
			Source(Supervised) & VGG-16 & 24.7 & 31.9 & 33.1 & 11.0 & 26.4 & 9.2 & 18.0 & 27.9 & 22.8 \\
			\hline
			DA Model$^{*}$ \cite{chen2018domain} & VGG-16 & 25.0 & 31.0 & 40.5 & 22.1 & 35.3 & 20.2 & 20.0 & 27.1 & 27.6 \\
			DA Model \cite{chen2018domain} & VGG-16 & 29.6 & 38.1 & 43.3 & 20.7 & 31.0 & 21.5 & 24.8 & 32.1 & 30.1 \\
			DT Model \cite{inoue2018cross} & VGG-16 & 25.4 & 39.3 & 42.4 & 24.9 & 40.4 & 23.1 & 25.9 & 30.4 & 31.5 \\
			SC-DA(Type3) \cite{zhu2019adapting} & VGG-16 & \textbf{33.5} & 38.0 & \textbf{48.5} & 26.5 & 39.0 & 23.3 & 28.0 & 33.6 & 33.8 \\
			SW-DA \cite{saito2019strong} & VGG-16 & 29.9 & 42.3 & 43.5 & 24.5 & 36.2 & 32.6 & 30.0 & 35.3 & 34.3 \\
			DD-MRL \cite{kim2019diversify} & VGG-16 & 30.8 & 40.5 & 44.3 & 27.2 & 38.4 & 34.5 & 28.4 & 32.2 & 34.6 \\
			MTOR \cite{cai2019exploring} & Resnet-50 & 30.6 & 41.4 & 44.0 & 21.9 & 38.6 & \textbf{40.6} & 28.3 & 35.6 & 35.1 \\
			\hline
			\textbf{Proposed(n=4)} & VGG-16 & \textbf{33.2} & \textbf{44.2} & 44.8 & \textbf{28.2} & \textbf{41.8} & 28.7 & \textbf{30.5} & \textbf{36.5} & \textbf{36.0} \\
			\hline
			Target(Supervised) & VGG-16 & 37.3 & 48.2 & 52.7 & 35.2 & 52.2 & 48.5 & 35.3 & 38.8 & 43.5 \\
			\hline
		\end{tabular}
	\end{center}
	\caption{Quantitative results on adaptation from \textit{Cityscapes} to \textit{Foggy Cityscapes}. The results of DA Model$^{*}$ is from its original paper and that of DA Model is implemented using our parameters. MTOR uses Resnet-50 as its backbone, while the others are VGG-16. Proposed(n) indicates that the model uses \textit{n} image domain classifiers.}
	\label{tab:foggy}
\end{table*}

In the real world, weather may change every day. It's critical that a detection model can perform consistently in different weather conditions. Therefore, we evaluate our model on \textit{Cityscapes} and \textit{Foggy Cityscapes} \cite{sakaridis2018semantic} datasets, which are used as source domain and target domain, respectively. The \textit{Foggy Cityscapes} dataset is rendered from \textit{Cityscapes} by adding fog noise, so it also has 2975 images in training set and 500 images in validation set. In this experiment, we report our results on all categories carried on the \textit{Foggy Cityscapes} validation set.

The results are summarized in Table \ref{tab:foggy}. The mAP of our method outperforms the baseline by +13.2\% and exceed all the other existing models. It is worth noting that the results of our method are only -7.4\% than the model supervised by target images. Among the performance of each category, our method performs as well as SC-DA(Type3) for person detection. We find the SC-DA(Type3) and MTOR model are very suited for car and train detection respectively, while the performances of other categories are greatly improved by our method.

\subsection{Synthetic Data Adaptation}

\begin{table}
	\begin{center}
		\begin{tabular}{|l|cccc|c|}
			\hline
			Method  & G & I & CTX & L & Car AP \\
			\hline
			Source(Supervised) & & & & & 34.3 \\
			\hline
			DA Model$^{*}$ \cite{chen2018domain} & $\checkmark$ & $\checkmark$ & & & 39.0 \\
			DA Model \cite{chen2018domain} & $\checkmark$ & $\checkmark$ & & & 39.4 \\
			SW-DA \cite{saito2019strong} & $\checkmark$ & & $\checkmark$ & $\checkmark$ & 40.1 \\
			SW-DA($\gamma=3$) \cite{saito2019strong} & $\checkmark$ & & $\checkmark$ & & 42.3 \\
			SC-DA(Type3) \cite{zhu2019adapting} & & & & & \textbf{43.0} \\
			\hline
			Proposed(n=3) & $\checkmark$ & $\checkmark$ & & $\checkmark$ & 42.3 \\
			Proposed(n=4) & $\checkmark$ & $\checkmark$ & & $\checkmark$ & 42.0 \\
			Proposed(n=5) & $\checkmark$ & $\checkmark$ & & $\checkmark$ & 42.7 \\
			Proposed(n=6) & $\checkmark$ & $\checkmark$ & & $\checkmark$ & \textbf{42.8} \\
			\hline
			Target(Supervised) & & & & & 62.7 \\
			\hline
		\end{tabular}
	\end{center}
	\caption{Results on adaptation from \textit{SIM 10k} to \textit{Cityscapes}. G, I, CTX, L indicate global alignment, instance-stage alignment, context-vector based regularization, and local alignment, respectively. DA Model$^{*}$ is from original paper and DA Model is implemented using our parameters.}
	\label{tab:sim10k}
\end{table}

We then show experiments about adaptation from synthetic images to real images. We utilized the \textit{SIM 10k} dataset as the synthetic source domain. This dataset contains 10000 images and 58701 bounding boxes of cars, which are collected from the video game Grand Theft Auto (GTA). All images are used in training. As for the target domain, we used \textit{Cityscapes} dataset. In addition, we only report the average precision of the cars on the validation set, since only the cars have annotations in \textit{SIM 10k}.

The results are summarized in Table \ref{tab:sim10k}. Specifically, compared with the baseline model which was supervised only on the source domain, the proposed model achieves +8.5\% performance gain using 6 domain classifiers. Compared with DA Model which doesn't use local alignment the proposed model achieves an improvement of +3.4\%. This indicates the importance of local alignment. SW-DA Model also adopts local alignment, but they only achieve +0.7\% performance gain, which suggests that the ability of one or two domain classifiers is limited. SC-DA(Type3) Model performs a little better than ours because their model is especially suitable for car detection as shown in the Adverse Weather Adaptation experiment. In addition, we find the performance can be further improved by increasing the number of domain classifiers from 3 to 6.

\begin{figure*}
	\begin{center}
		\includegraphics[width=0.85\linewidth]{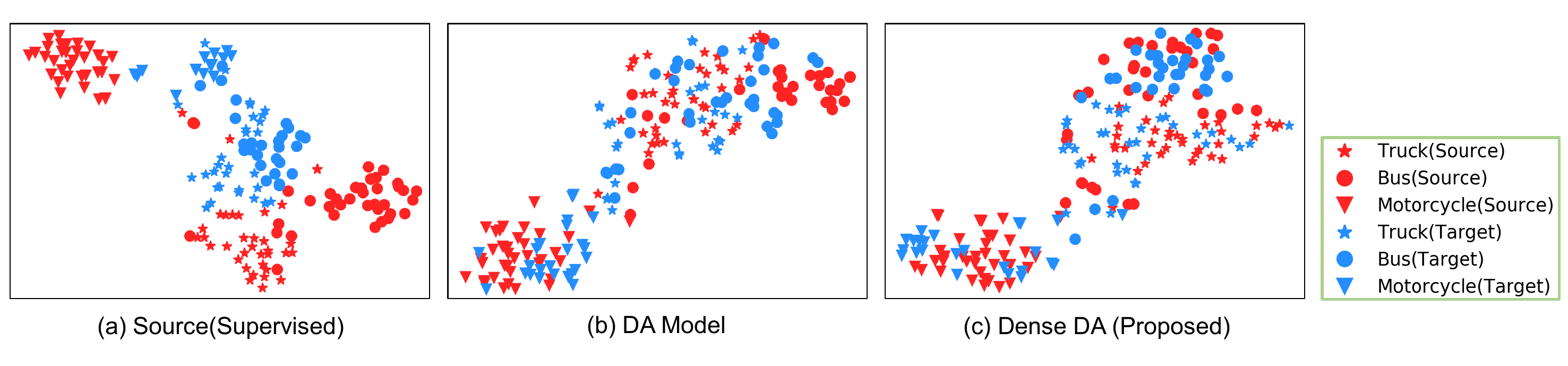}
	\end{center}
	\caption{Visualization of instance (ROI) features using the t-SNE algorithm: (a) Features from the baseline model (supervised learning by data from source domain) (b) Features from the DA model (c) Features from our multi-level DA model. The color of points represents the domain label and each shape indicates a category of this instance, best viewed in color.}
	\label{fig:ins_vis}
\end{figure*}

\begin{figure*}
	\begin{center}
		\includegraphics[width=0.8\linewidth]{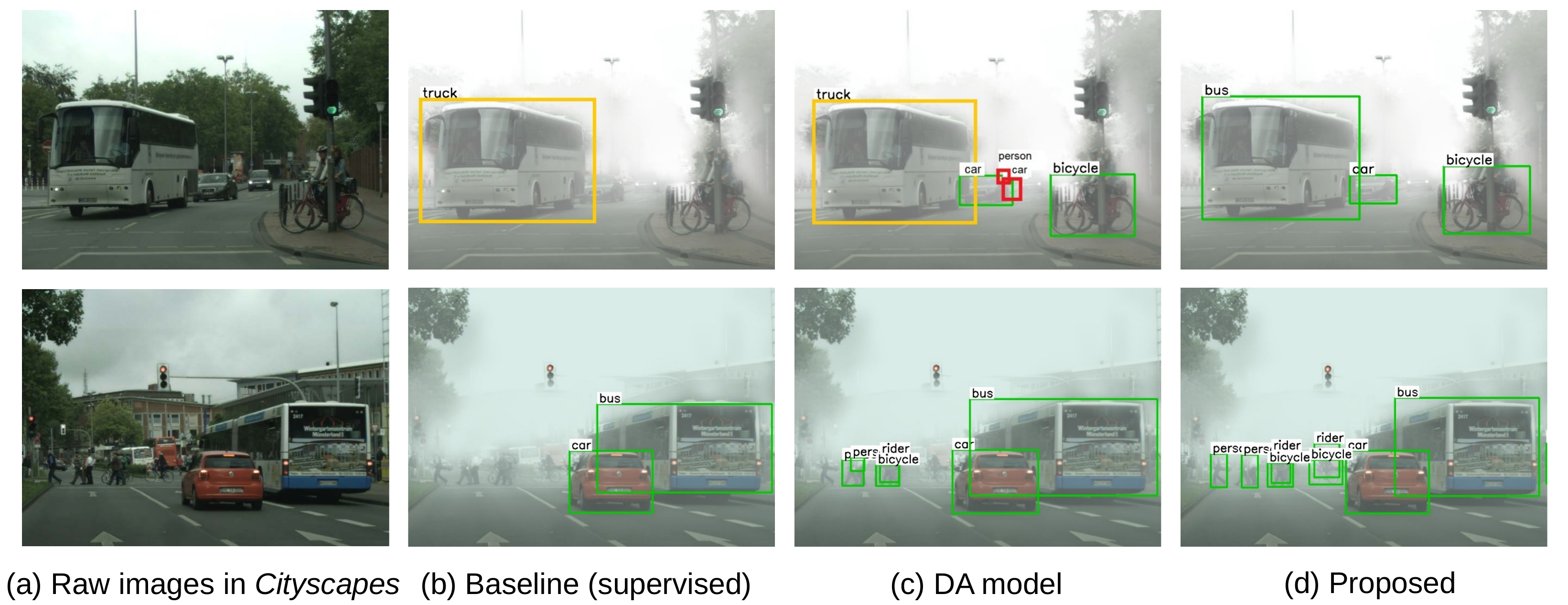}
	\end{center}
	\caption{Qualitative results on adaptation from \textit{Cityscapes} to \textit{Foggy Cityscapes} dataset. (a) Raw images in \textit{Cityscapes}, (b) Baseline model (supervised on \textit{Cityscapes}), (c) DA model, (d) Proposed model (n=4). The raw images in \textit{Cityscapes} are for reading only. Boxes with green color mean correct results, red color denotes false positives and yellow color means misclassification.}
	\label{fig:foggy}
\end{figure*}

\subsection{Cross Camera Adaptation}

\begin{table*}
	\begin{center}
		\begin{tabular}{|l|ccccc|c|}
			\hline
			Method  & Person & Rider & Car & Truck & Train & Mean AP \\
			\hline
			Source(Supervised) & 47.8 & 22.0 & \textbf{75.2} & 12.4 & 12.6 & 34.0 \\
			\hline
			DA Model \cite{chen2018domain} & 40.9 & 16.1 & 70.3 & 23.6 & 21.2 & 34.4 \\
			\hline
			\textbf{Proposed(n=4)} & \textbf{53.0} & \textbf{24.5} & 72.2 & \textbf{28.7} & \textbf{25.3} & \textbf{40.7}  \\
			\hline
		\end{tabular}
	\end{center}
	\caption{Quantitative results on adaptation from \textit{Cityscapes} to \textit{KITTI}. Since the detection objects are changed, we only give the results of DA Model which is implemented using our parameters.}
	\label{tab:kitti}
\end{table*}

In this experiment, we aim to analyze the adaptation for the images under different camera setups. We utilize the \textit{Cityscapes} dataset as the source domain and \textit{KITTI} dataset as the target domain. The \textit{KITTI} dataset consists of 7481 images, which have original resolution of 1250x375. They are resized so that the shorter length is 600 pixels long. In addition, the \textit{KITTI} dataset is used in both adaptation and evaluation.

We report the mAP of 5 categories with a threshold of 0.5. However, we find the classification standard of categories in two domain datasets is different. So we classify 'Car' and 'Van' as 'Car', 'Person' and 'Person sitting' as 'Person', then we convert 'Tram' to 'Train', 'Cyclist' to 'Rider' in the \textit{KITTI} dataset, which is different from \cite{chen2018domain}.

The results are summarized in Table \ref{tab:kitti}. In our experimental settings, the baseline model already has a good ability for person, rider, and car detection because both the source and target domain datasets are from real world and the domain shift in these three categories is very small. We find the introduction of domain classifier caused a performance drop. However, our method not only reduces the bad influence in car detection but also greatly improve the detections of persons and riders. As for other categories, both DA model and our method perform better than the baseline but ours goes far beyond the other two. The results indicate our method can achieve a better performance when the domain shift is large, and reduce the possible instability caused by domain adaptation when the domain shift is very small.

\subsection{Analysis}

\paragraph{Visualization of image-stage features}
We visualize the image-stage features using the t-SNE algorithm \cite{maaten2008visualizing} in Figure \ref{fig:img_vis}. All samples are from validation set of \textit{Foggy Cityscapes} dataset. The global features are aligned well in both DA model and the proposed Multi-level DA model. However, the local features between source and target domain are mismatched in DA model. This result confirms the first limitation of DA model in Section~\ref{sec.method}. Our method can align local features more effectively, which benefits from the proposed strategy of multi-level adaptation.

\paragraph{Visualization of instance-stage features}
We extract the features of several region proposals (before the final classification and regression layer). The t-SNE embedding of these features (from \textit{Foggy Cityscapes} dataset) is shown in Figure \ref{fig:ins_vis}. Notice the truck (star) and bus (circle) in Figure \ref{fig:ins_vis}(b). Although the DA model can align the marginal distribution to some extent, the categories are not discriminated well. But our model can align distribution and discriminate categories better. Such improvement explains why our model outperforms the baseline and the DA model.

\paragraph{Qualitative examples of detection results}
Figure \ref{fig:foggy} shows some typical detection results. In the first row, the baseline model almost ignored all objects. The DA model successfully detects a car and a bicycle, but it incorrectly classifies the bus as a truck, and has some false positives. Our model correctly detected the bus. In the second row, our model correctly detects more persons and bicycles in fog, even if recognizing them is challenging for humans.

\section{Conclusion}
In this paper, we propose an effective approach for cross-domain object detection. We introduce multiple domain classifiers to enforce multi-level adversarial training to improve the overall feature alignment. The proposed method outperforms the existing methods in several experiments. Moreover, the visualizations of feature distributions prove that our model can get more effective alignment than other models. However, the implementation of adversarial training in our model is based on gradient reversal layers (GRLs), which may cause instability in training. In future work, we plan to further investigate how to improve the accuracy and robustness of our models.

\paragraph{Acknowledgments.} This work was supported by National Key R\&D Program of China (No. 2018YFC0910\-700), NSFC (81801778, 11831002, 61625201 and 6152\-7804), Beijing Natural Science Foundation (Z180001) and Qualcomm University Research Grant.
{\small
\bibliographystyle{ieee}
\bibliography{egbib}

\begin{thebibliography}{10}\itemsep=-1pt

\bibitem{cai2019exploring}
Q.~Cai, Y.~Pan, C.-W. Ngo, X.~Tian, L.~Duan, and T.~Yao.
\newblock Exploring object relation in mean teacher for cross-domain detection.
\newblock In {\em Proceedings of the IEEE Conference on Computer Vision and
  Pattern Recognition}, pages 11457--11466, 2019.

\bibitem{chen2018domain}
Y.~Chen, W.~Li, C.~Sakaridis, D.~Dai, and L.~Van~Gool.
\newblock Domain adaptive faster r-cnn for object detection in the wild.
\newblock In {\em Proceedings of the IEEE Conference on Computer Vision and
  Pattern Recognition}, pages 3339--3348, 2018.

\bibitem{cordts2016cityscapes}
M.~Cordts, M.~Omran, S.~Ramos, T.~Rehfeld, M.~Enzweiler, R.~Benenson,
  U.~Franke, S.~Roth, and B.~Schiele.
\newblock The cityscapes dataset for semantic urban scene understanding.
\newblock In {\em Proceedings of the IEEE conference on computer vision and
  pattern recognition}, pages 3213--3223, 2016.

\bibitem{dai2016r}
J.~Dai, Y.~Li, K.~He, and J.~Sun.
\newblock R-fcn: Object detection via region-based fully convolutional
  networks.
\newblock In {\em Advances in neural information processing systems}, pages
  379--387, 2016.

\bibitem{ganin2015unsupervised}
Y.~Ganin and V.~Lempitsky.
\newblock Unsupervised domain adaptation by backpropagation.
\newblock In {\em International Conference on Machine Learning}, pages
  1180--1189, 2015.

\bibitem{ganin2016domain}
Y.~Ganin, E.~Ustinova, H.~Ajakan, P.~Germain, H.~Larochelle, F.~Laviolette,
  M.~Marchand, and V.~Lempitsky.
\newblock Domain-adversarial training of neural networks.
\newblock {\em The Journal of Machine Learning Research}, 17(1):2096--2030,
  2016.

\bibitem{geiger2013vision}
A.~Geiger, P.~Lenz, C.~Stiller, and R.~Urtasun.
\newblock Vision meets robotics: The kitti dataset.
\newblock {\em The International Journal of Robotics Research},
  32(11):1231--1237, 2013.

\bibitem{he2017mask}
K.~He, G.~Gkioxari, P.~Doll{\'a}r, and R.~Girshick.
\newblock Mask r-cnn.
\newblock In {\em Proceedings of the IEEE international conference on computer
  vision}, pages 2961--2969, 2017.

\bibitem{hoffman2017cycada}
J.~Hoffman, E.~Tzeng, T.~Park, J.-Y. Zhu, P.~Isola, K.~Saenko, A.~A. Efros, and
  T.~Darrell.
\newblock Cycada: Cycle-consistent adversarial domain adaptation.
\newblock {\em arXiv preprint arXiv:1711.03213}, 2017.

\bibitem{huang2018domain}
H.~Huang, Q.~Huang, and P.~Krahenbuhl.
\newblock Domain transfer through deep activation matching.
\newblock In {\em Proceedings of the European Conference on Computer Vision
  (ECCV)}, pages 590--605, 2018.

\bibitem{inoue2018cross}
N.~Inoue, R.~Furuta, T.~Yamasaki, and K.~Aizawa.
\newblock Cross-domain weakly-supervised object detection through progressive
  domain adaptation.
\newblock In {\em Proceedings of the IEEE Conference on Computer Vision and
  Pattern Recognition}, pages 5001--5009, 2018.

\bibitem{johnson2017driving}
M.~Johnson-Roberson, C.~Barto, R.~Mehta, S.~N. Sridhar, K.~Rosaen, and
  R.~Vasudevan.
\newblock Driving in the matrix: Can virtual worlds replace human-generated
  annotations for real world tasks?
\newblock In {\em 2017 IEEE International Conference on Robotics and Automation
  (ICRA)}, pages 746--753. IEEE, 2017.

\bibitem{kim2019diversify}
T.~Kim, M.~Jeong, S.~Kim, S.~Choi, and C.~Kim.
\newblock Diversify and match: A domain adaptive representation learning
  paradigm for object detection.
\newblock In {\em Proceedings of the IEEE Conference on Computer Vision and
  Pattern Recognition}, pages 12456--12465, 2019.

\bibitem{liu2017unsupervised}
M.-Y. Liu, T.~Breuel, and J.~Kautz.
\newblock Unsupervised image-to-image translation networks.
\newblock In {\em Advances in Neural Information Processing Systems}, pages
  700--708, 2017.

\bibitem{long2015learning}
M.~Long, Y.~Cao, J.~Wang, and M.~I. Jordan.
\newblock Learning transferable features with deep adaptation networks.
\newblock {\em international conference on machine learning}, pages 97--105,
  2015.

\bibitem{maaten2008visualizing}
L.~v.~d. Maaten and G.~Hinton.
\newblock Visualizing data using t-sne.
\newblock {\em Journal of machine learning research}, 9(Nov):2579--2605, 2008.

\bibitem{mao2017least}
X.~Mao, Q.~Li, H.~Xie, R.~Y. Lau, Z.~Wang, and S.~Paul~Smolley.
\newblock Least squares generative adversarial networks.
\newblock In {\em Proceedings of the IEEE International Conference on Computer
  Vision}, pages 2794--2802, 2017.

\bibitem{redmon2018yolov3}
J.~Redmon and A.~Farhadi.
\newblock Yolov3: An incremental improvement.
\newblock {\em arXiv preprint arXiv:1804.02767}, 2018.

\bibitem{ren2015faster}
S.~Ren, K.~He, R.~Girshick, and J.~Sun.
\newblock Faster r-cnn: Towards real-time object detection with region proposal
  networks.
\newblock In {\em Advances in neural information processing systems}, pages
  91--99, 2015.

\bibitem{ren2018cross}
Z.~Ren and Y.~Jae~Lee.
\newblock Cross-domain self-supervised multi-task feature learning using
  synthetic imagery.
\newblock In {\em Proceedings of the IEEE Conference on Computer Vision and
  Pattern Recognition}, pages 762--771, 2018.

\bibitem{saito2019strong}
K.~Saito, Y.~Ushiku, T.~Harada, and K.~Saenko.
\newblock Strong-weak distribution alignment for adaptive object detection.
\newblock In {\em Proceedings of the IEEE Conference on Computer Vision and
  Pattern Recognition}, pages 6956--6965, 2019.

\bibitem{saito2018maximum}
K.~Saito, K.~Watanabe, Y.~Ushiku, and T.~Harada.
\newblock Maximum classifier discrepancy for unsupervised domain adaptation.
\newblock In {\em Proceedings of the IEEE Conference on Computer Vision and
  Pattern Recognition}, pages 3723--3732, 2018.

\bibitem{sakaridis2018semantic}
C.~Sakaridis, D.~Dai, and L.~Van~Gool.
\newblock Semantic foggy scene understanding with synthetic data.
\newblock {\em International Journal of Computer Vision}, pages 1--20, 2018.

\bibitem{simonyan2014very}
K.~Simonyan and A.~Zisserman.
\newblock Very deep convolutional networks for large-scale image recognition.
\newblock {\em arXiv preprint arXiv:1409.1556}, 2014.

\bibitem{tzeng2017adversarial}
E.~Tzeng, J.~Hoffman, K.~Saenko, and T.~Darrell.
\newblock Adversarial discriminative domain adaptation.
\newblock In {\em Proceedings of the IEEE Conference on Computer Vision and
  Pattern Recognition}, pages 7167--7176, 2017.

\bibitem{Tzeng_2017_CVPR}
E.~Tzeng, J.~Hoffman, K.~Saenko, and T.~Darrell.
\newblock Adversarial discriminative domain adaptation.
\newblock In {\em The IEEE Conference on Computer Vision and Pattern
  Recognition (CVPR)}, July 2017.

\bibitem{tzeng2014deep}
E.~Tzeng, J.~Hoffman, N.~Zhang, K.~Saenko, and T.~Darrell.
\newblock Deep domain confusion: Maximizing for domain invariance.
\newblock {\em arXiv preprint arXiv:1412.3474}, 2014.

\bibitem{zhang2018collaborative}
W.~Zhang, W.~Ouyang, W.~Li, and D.~Xu.
\newblock Collaborative and adversarial network for unsupervised domain
  adaptation.
\newblock In {\em Proceedings of the IEEE Conference on Computer Vision and
  Pattern Recognition}, pages 3801--3809, 2018.

\bibitem{zhu2019adapting}
X.~Zhu, J.~Pang, C.~Yang, J.~Shi, and D.~Lin.
\newblock Adapting object detectors via selective cross-domain alignment.
\newblock In {\em Proceedings of the IEEE Conference on Computer Vision and
  Pattern Recognition}, pages 687--696, 2019.

\end{thebibliography}
}

\end{document}